\documentclass[conference]{IEEEtran}
\usepackage{graphicx} 
\usepackage[T1]{fontenc}
\usepackage{tabularx}
\usepackage{amsmath}

\title{VRScout: Towards Real-Time, Autonomous Testing of Virtual Reality Games}
\author{
    \IEEEauthorblockN{Yurun Wu, Yousong Sun, Burkhard Wunsche}
    \IEEEauthorblockA{School of Computer Science\\
                      University of Auckland\\
            yurun.wu, yousong.sun, burkhard@auckland.ac.nz}
    \and
    \IEEEauthorblockN{Jia Wang}
    \IEEEauthorblockA{School of Advanced Technology\\
                      Xi'an Jiaotong-Liverpool University\\
            Jia.Wang02@xjtlu.edu.cn}
    \and
    \IEEEauthorblockN{Elliott Wen}
    \IEEEauthorblockA{School of Computer Science\\
                      University of Auckland\\
            elliott.wen@auckland.ac.nz}
}

\begin{document}

\maketitle
\begin{abstract}
Virtual Reality (VR) has rapidly become a mainstream platform for gaming and interactive experiences, yet ensuring the quality, safety, and appropriateness of VR content remains a pressing challenge. Traditional human-based quality assurance is labor-intensive and cannot scale with the industry’s rapid growth. While automated testing has been applied to traditional 2D and 3D games, extending it to VR introduces unique difficulties due to high-dimensional sensory inputs and strict real-time performance requirements.
We present VRScout, a deep learning–based agent capable of autonomously navigating VR environments and interacting with virtual objects in a human-like and real-time manner. VRScout learns from human demonstrations using an enhanced Action Chunking Transformer that predicts multi-step action sequences. This enables our agent to capture higher-level strategies and generalize across diverse environments. To balance responsiveness and precision, we introduce a dynamically adjustable sliding horizon that adapts the agent’s temporal context at runtime.
We evaluate VRScout on commercial VR titles and show that it achieves expert-level performance with only limited training data, while maintaining real-time inference at 60 FPS on consumer-grade hardware. These results position VRScout as a practical and scalable framework for automated VR game testing, with direct applications in both quality assurance and safety auditing.

\end{abstract}
\section{Introduction}
Virtual Reality (VR) has rapidly evolved into a mainstream platform for gaming and interactive experiences, with the global VR gaming market projected to reach \$84 billion by 2028. However, reports of physical accidents and inappropriate content underscore the need for broader testing to ensure the quality, safety, and suitability of VR games.
A core challenge in VR testing is thoroughly navigating each scenario and interacting with in-game objects. Traditionally, this work has relied on human quality assurance testers, but the approach is labor-intensive, scales poorly with the industry’s rapid growth. It also raises ethical concerns from the potential exposure of testers to harmful conditions.

Recently, some researchers have explored the use of AI for testing traditional 2D and 3D games automatically~\cite{paduraru2022rivergame,rege2024level,roque2025literature,zhang2025human}. Nevertheless, relatively little work has focused specifically on VR games. This gap exists because automated testing in VR is significantly more challenging due to high-dimensional inputs, such as immersive 360-degree first-person images, three-dimensional head and hand tracking, and multiple controller buttons. They greatly expand the state space an agent must manage. Moreover, real-time performance requirements add further complexity, as agents must operate at high frame rates to keep pace with the VR simulation.

In this paper, we introduce VRScout, a deep learning powered agent capable of autonomously navigating VR environments and interacting with virtual objects in a human-like and real-time manner. It paves the way for automated testing to detect implementation bugs and identify inappropriate content.
VRScout learns from human demonstrations to act naturally and efficiently through an Action Chunking Transformer (ACT)~\cite{zhao2023learning,buamanee2024bi}. It processes a time series of VR scene images and predicts multi-step sequences of controller movements and button actions (e.g., move forward, then turn right, and press button). Compared to single-step prediction, this approach captures higher-level, longer-term, and noise-resistant agent strategies that generalize across environments. 
We introduce an optimization to adapt the ACT for VR environments. We propose a dynamically adjustable sliding horizon, defined as the number of consecutive VR scene images the agent processes before committing to an action sequence. By automatically adjusting the prediction horizon at runtime, VRScout can balance faster inference speeds (shorter horizons) with higher action accuracy (longer horizons), depending on the VR game.

We implement and evaluate VRScout on three popular commercial VR games including \textit{Beat Saber}\footnote{https://beatsaber.com/}, \textit{SuperHot}\footnote{https://superhotgame.com/}, and \textit{Pistol Whip}\footnote{https://www.meta.com/en-gb/experiences/pistol-whip/2104963472963790/}.
Our experiments demonstrate that VRScout possesses two major advantages. Firstly, VRScout requires only a small amount of training data. For example, in \textit{Beat Saber}, four hours of human expert demonstration are sufficient for it to achieve expert-level performance. It demonstrates VRScout’s data-efficient learning capability. Secondly, VRScout achieves real-time inference at 60 FPS when running on a consumer-grade NVIDIA 4090. This matches the typical frame rates of VR games. Its performance can be further enhanced through optimization techniques such as quantization and model compilation. These results highlight VRScout as a practical and scalable solution for automated VR testing.

In the following sections, we first review related work in Section~\ref{related}. We then describe the architecture of VRScout in Section~\ref{method}, detail its learning and decision-making mechanisms, and present extensive evaluations across multiple commercial VR games in Section~\ref{experiment}. To support research reproducibility, we provide open-source access to our system and dataset via GitHub.

\section{Related Works}\label{related}

\noindent \textbf{Automated Game Testing:} AI-driven automated testing in 2D and 3D games has emerged as an active area of research. Early research primarily relied on Reinforcement Learning (RL), where agents are trained to explore game environments to identify bugs, exploits, or gameplay imbalances~\cite{berner2019dota, bergdahl2020augmenting, zheng2019wuji, ferdous2022towards}. However, these RL-based approaches often require careful design of reward functions, which can be time-consuming and may not generalize well across different games~\cite{gillberg2023technical}. More recent work~\cite{zheng2022imitation,sestini2023towards, schafer2023visual} has explored Imitation Learning (IL) as an alternative. By leveraging demonstrations from human players or expert agents, IL agents can autonomously explore and interact with virtual environments in a human-like manner, without manually specifying complex reward functions. 

Imitation learning (IL) agents can be trained using various approaches. For example, Behavior Cloning~\cite{torabi2018behavioral,kanervisto2020playing,kanervisto2020benchmarking, pearce2022counter} enables agents to learn a mapping from game states to corresponding actions. Inverse Reinforcement Learning (IRL)~\cite{osa2018algorithmic}, in contrast, infers a reward function from expert demonstrations and trains the agent to optimize this reward. A more advanced approach is Generative Adversarial Imitation Learning (GAIL)~\cite{ho2016generative}, which integrates IRL with adversarial learning. In GAIL, a generator is trained to imitate expert actions, while a discriminator distinguishes between real and generated behaviors. This encourages agents to perform more complex navigation tasks.
In this work, we adapt the ACT. Unlike previous approaches, ACT models sequences of actions in temporal chunks. This allows it to capture long-range dependencies and to efficiently integrate visual observations from the game environment.

\noindent \textbf {Automated VR Game Testing:} 
Despite significant progress in automated game testing, relatively little research has focused specifically on VR games. This gap arises because VR testing presents unique challenges due to its large state-action space, which encompasses continuous 6-DoF tracking of the head and hands, 360-degree visual input, and a variety of controller interactions~\cite{roque2025literature}. Moreover, agents must operate in real time at high frame rates (e.g., 60 Hz) to remain synchronized with the VR simulation. This imposes strict constraints on model inference latency.

The work most closely related to this paper is by Qin et al.~\cite{qin2024utilizing}, who investigated VR exploration testing within a purpose-built Unity experimental environment. In their study, ChatGPT is tasked with identifying the positions of objects within the field of view and drawing outlines such as bounding boxes, as well as approaching virtual objects within the scene. However, this prior work primarily focuses on investigating AI models' capacity to understand the logical and causal relationships of VR objects in a controlled environment. In contrast, our approach directly interacts with off-the-shelf VR games, operating in fully unmodified environments and achieving expert-level gameplay performance. This work serves as an initial step toward automated testing in VR, highlighting the potential for more practical and scalable evaluation methods.

\section{Methodology}\label{method}
\begin{figure*}[t]
    \centering
    \includegraphics[width=0.9\textwidth]{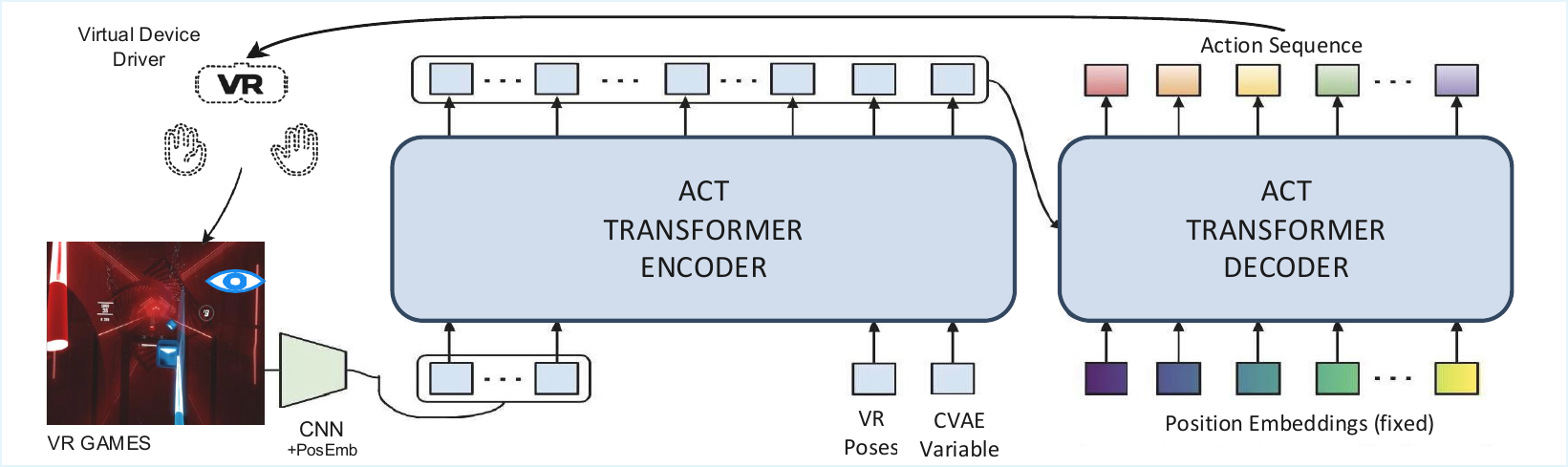}
    \caption{Architecture of the VR agent. }
    \label{fig:workflow}
\end{figure*}

Figure~\ref{fig:workflow} illustrates the workflow of our system. VR scene images and user actions are first encoded into feature vectors. These features are then grouped into short temporal chunks, which are fed into an ACT model encoder. The encoder employs a transformer architecture to generate context-aware representations. The decoder uses these representations, together with its previous outputs, to predict the next chunk of actions. These predicted actions are injected into the games via virtual VR controllers to produce the subsequent game states and continue gameplay.

\noindent\textbf{Model Input and Feature Extraction:}
Our model takes two types of inputs: VR images and user actions. VR images are sampled at 30 Hz and processed by a \textit{ResNet-18}~\cite{he2016deep} encoder pretrained on ImageNet to produce feature vectors. User actions are encoded into a separate feature vector, which includes position and quaternion data for three devices (headset, left controller, and right controller), four values for controller triggers and grips, four joystick axis values (x and y for each controller), the position of the VR origin, and the states of relevant buttons. In addition, we incorporate a CVAE-style latent embedding~\cite{zhao2023learning, sohn2015learning} conditioned on the current observation and the target action chunk. This latent vector captures variability in human demonstrations and serves as an enhanced feature representation, enabling the model to generate consistent yet flexible action predictions.

\begin{figure}[h]
    \centering
    \includegraphics[width=0.5\textwidth]{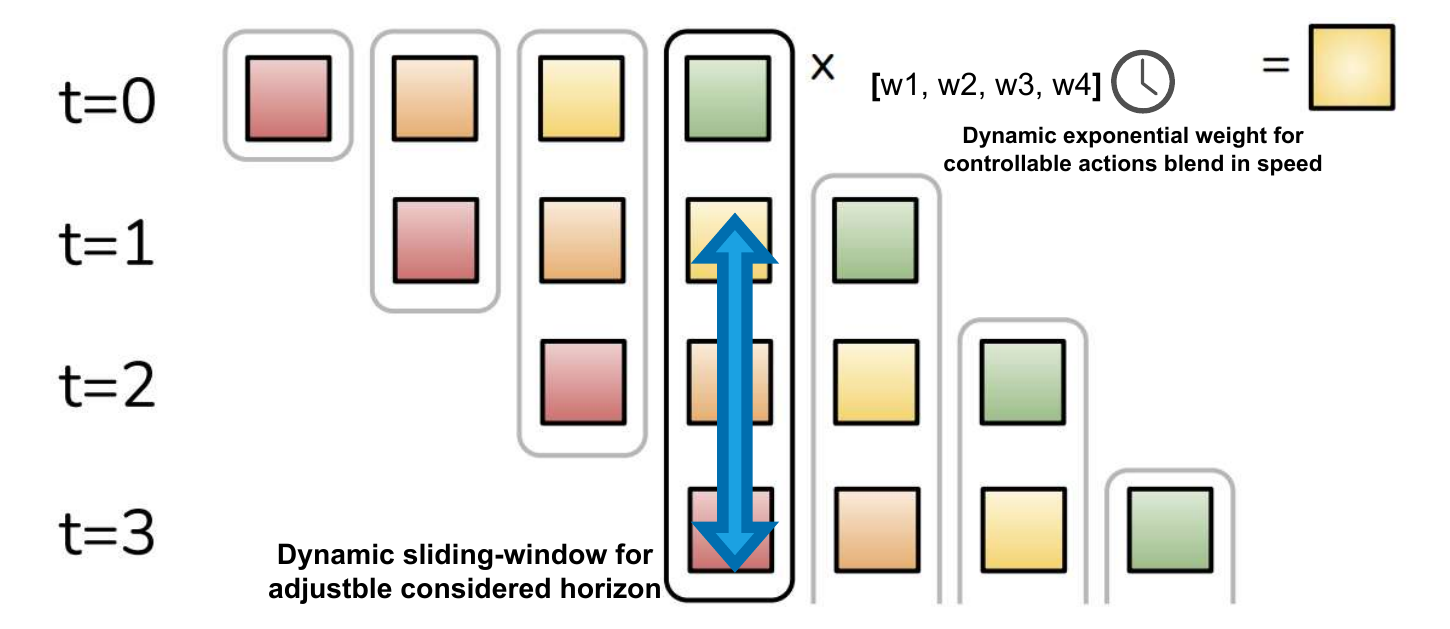}
    \caption{Temporal Ensembling with Dynamic Sliding Window}
    \label{fig:dynamicwindow}
\end{figure}

\noindent\textbf{Temporal Ensembling and Dynamic Sliding Window:}
In ACT, the prediction horizon refers to the number of future actions generated in a single model call. A shorter horizon enables faster inference, while a longer horizon improves prediction quality by producing smoother and more temporally consistent actions. The optimal horizon often depends on the game: fast-paced titles typically require shorter horizons, whereas slower-paced games can benefit from longer ones.

One approach to balancing these trade-offs is \emph{temporal ensembling}, which combines the most recent action prediction with past predictions of the same action from earlier model calls. The merging employs exponential weighting, where each past prediction is assigned a weight as follows:
\[
w_i = \exp(-m \cdot i).
\]
This strategy reduces noise and produces smoother, more temporally consistent actions, while still allowing the system to respond promptly to new inputs. By adjusting the weighting, it can flexibly balance reactivity and stability depending on the task~\cite{zhao2023learning}.

Building on this idea, we further adopt a more advanced \emph{dynamic sliding-window} approach to go beyond tuning a single parameter. Figure~\ref{fig:dynamicwindow} illustrates the action sequences predicted over the time steps $t=0$ to $t=3$. The blue arrow denotes the current window length, which is dynamically tuned to determine how many past predictions contribute to the final action. In addition, a dynamic exponential weighting function governs the relative influence of these past predictions, so that more recent outputs can dominate while older ones gradually decay in importance, allowing it to control how quickly new predictions are blended with past actions during inference. This design lets the system adapt both the effective prediction horizon and the influence of historical context, allowing the agent to flexibly balance reactivity and stability in its actions depending on the task. 
Our approach uses feedback signals to automatically adjust the horizon length and the exponential weight.  
More specifically, 
we evaluate multiple factors such as the average motion speed over short horizons. prediction entropy, action variance, and historical consistency in Section~\ref{experiment}.

\noindent\textbf{Model Output:}
For each model call, the network outputs a sequence of absolute positions and quaternion rotations for each tracked device, post-processed with the adaptive temporal ensembling described above to obtain the action for the current timestep. These actions are applied directly through a virtual VR device driver, allowing the agent to interact with commercial VR games without software modification. The virtual VR device driver is developed upon the Valve OpenVR SDK and exposes a virtual headset together with two virtual controllers inside SteamVR. By injecting the model’s predictions as device poses and button states, the driver enables interaction with VR titles on the Steam platform as if a real human player were operating physical devices. To ensure valid and stable orientations, quaternions are normalized post-prediction to maintain unit length.

\noindent\textbf{Model Training and Loss Function:}
Our training objective combines continuous pose regression, discrete button prediction, and latent regularization. The final loss for the policy is defined as
\begin{equation}
\mathcal{L} = L_{\mathrm{cont\_L1}} + 0.2 \, L_{\mathrm{bool}} + \lambda_{\mathrm{KL}} \, \mathrm{KL}\big(q(z|x) \,\|\, p(z)\big),
\end{equation}
where $L_{\mathrm{cont_L1}}$ denotes the mean L1 error (MAE) across continuous pose outputs, $L_{\mathrm{bool}}$ corresponds to the loss on Boolean states (e.g., trigger, grip) and is defined as a weighted combination of binary cross-entropy (70\%) and L1 loss (30\%), and the final term is the Kullback–Leibler divergence over the latent distribution of the CVAE. 
The weighting of $L_{\mathrm{bool}}$ leverages the strengths of both binary cross-entropy and L1 loss. Cross-entropy drives accurate classification by penalizing confident errors, while L1 provides a smoother signal that stabilizes predictions near the decision boundary. Using a 0.7/0.3 balance ensures the agent remains decisive without becoming overconfident, leading to steadier, more human-like control of discrete VR actions such as trigger pulls or grip squeezes. This reduces jitter and accidental activations during fast gameplay, improving agent reliability.

\section{Experiment Results}\label{experiment}

In this section, we describe our experimental setup and present our system's performance on off-the-shelf games.

\subsection{Experiment Settings:} 

\noindent\textbf{Data Collection and Setup:}
Meta Quest 3 was the VR device used to collect expert gameplay demonstrations. All the games were run on Valve’s Steam platform with SteamVR. We extended the CLOVR~\cite{Martinez2024CLOVR} and VRHook~\cite{wen2022vrhook} to realise synchronised collection of VR scene images and actions at 30Hz. The dataset was obtained from a single human player’s gameplay, consisting of 3.5 hours of demonstrations for each game. The player had more than 200 hours of experience in Beat Saber and was therefore able to provide expert-level demonstrations in that title. By contrast, their proficiency in Pistol Whip was more limited, and they were a beginner in SuperHotVR.

\noindent\textbf{Model Training:}
For training, we used an NVIDIA A100 with PyTorch and CUDA acceleration. Training a model typically required about five hours of wall-clock time. The model was trained with the Adam optimizer using a learning rate of 3e-5 and a batch size of 32. The dataset was split 80/20 for training and validation.

It should be noted that the quality of training strongly depends on how demonstration data is organized, we investigated the impact of different sampling strategies.
Specifically, we compared fully randomised sampling, where batches are drawn from any part of the dataset, with episodic randomised sampling, where each batch spans multiple demonstration episodes to ensure diversity.  

Fully randomised sampling can repeatedly select samples from the same demonstration episode, which may make the model's task easier in the short term but can reduce the effective learning rate and limit generalisation. In contrast, episodic sampling consistently outperformed fully randomised sampling in terms of in-game score and generalisability. In Beat Saber, for example, episodic sampling produced on average a 8\% higher score when tested on five maps that ranged from easier to expert levels. This highlights the importance of balanced episode coverage for stable imitation learning in VR gameplay.

\begin{figure}[h]
\centering
\includegraphics[width=0.9\linewidth]{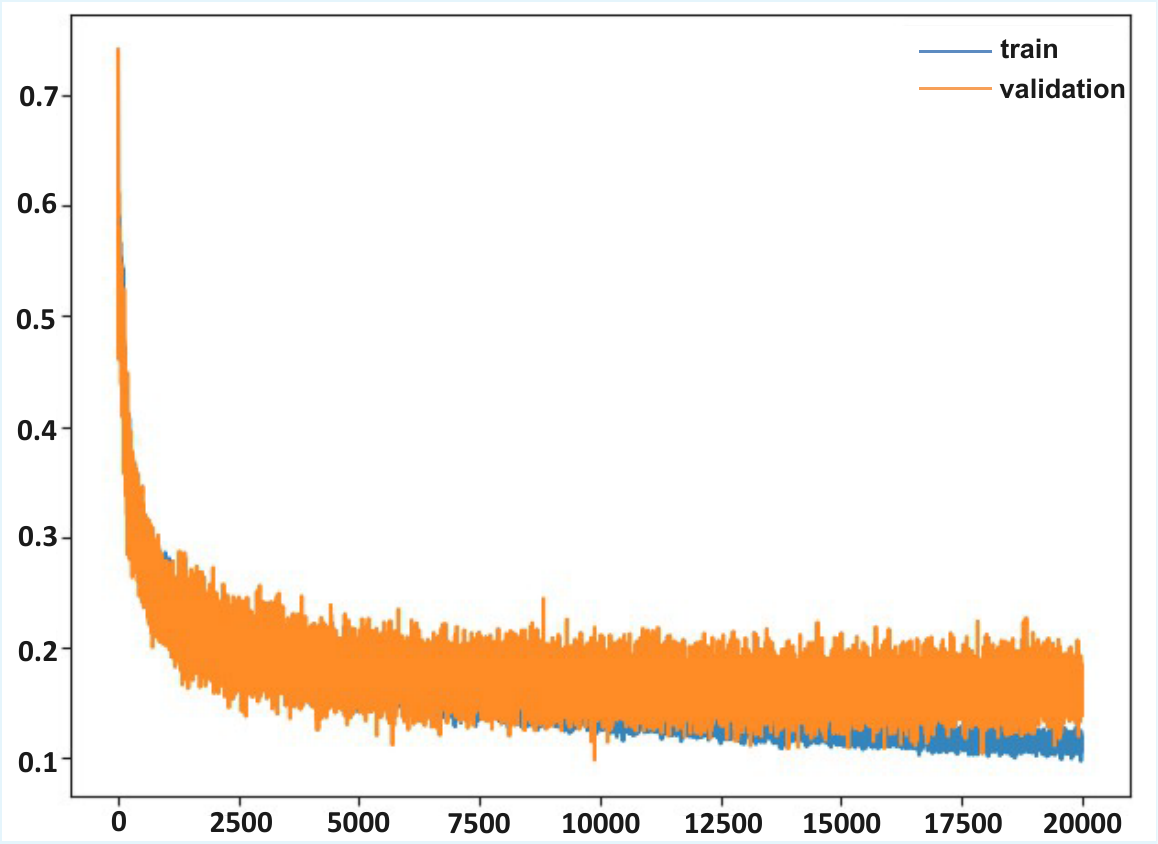}
\caption{Training curve plot (L1 loss vs. iterations).}
\label{fig:training_curve_l1}
\end{figure}

For Beat Saber, our best performing ACT model achieved an L1 loss of 0.098 and a total validation loss of 0.1. Figure~\ref{fig:training_curve_l1} shows the training curve of L1 loss over iterations, indicating that the model converges stably while generalising well to unseen observations. We also evaluated the agent’s in-game performance at 5000, 10000, 15000, and 20000 iterations. The improvement slowed after 10000 iterations, with the validation loss stabilising beyond this point. Training was capped at 20000 iterations, by which stage the dataset likely achieved near-complete coverage through episodic sampling. Here, coverage is defined as the percentage of samples that are seen at least once by the model during training. Total training time for this schedule was about five hours on an A100.

\subsection{Model  Performance} 
Inference was tested on both RTX 4090 and RTX 3060 GPUs. On the RTX 4090, the agent maintained a stable in-game frame rate of 60 Hz with full action aggregation (smoothing consecutive predictions). On the RTX 3060, 35 Hz was achieved by reducing smoothing, showing the model remains deployable even on mid-tier hardware.

\begin{figure}[h]
\centering
\includegraphics[width=\linewidth]{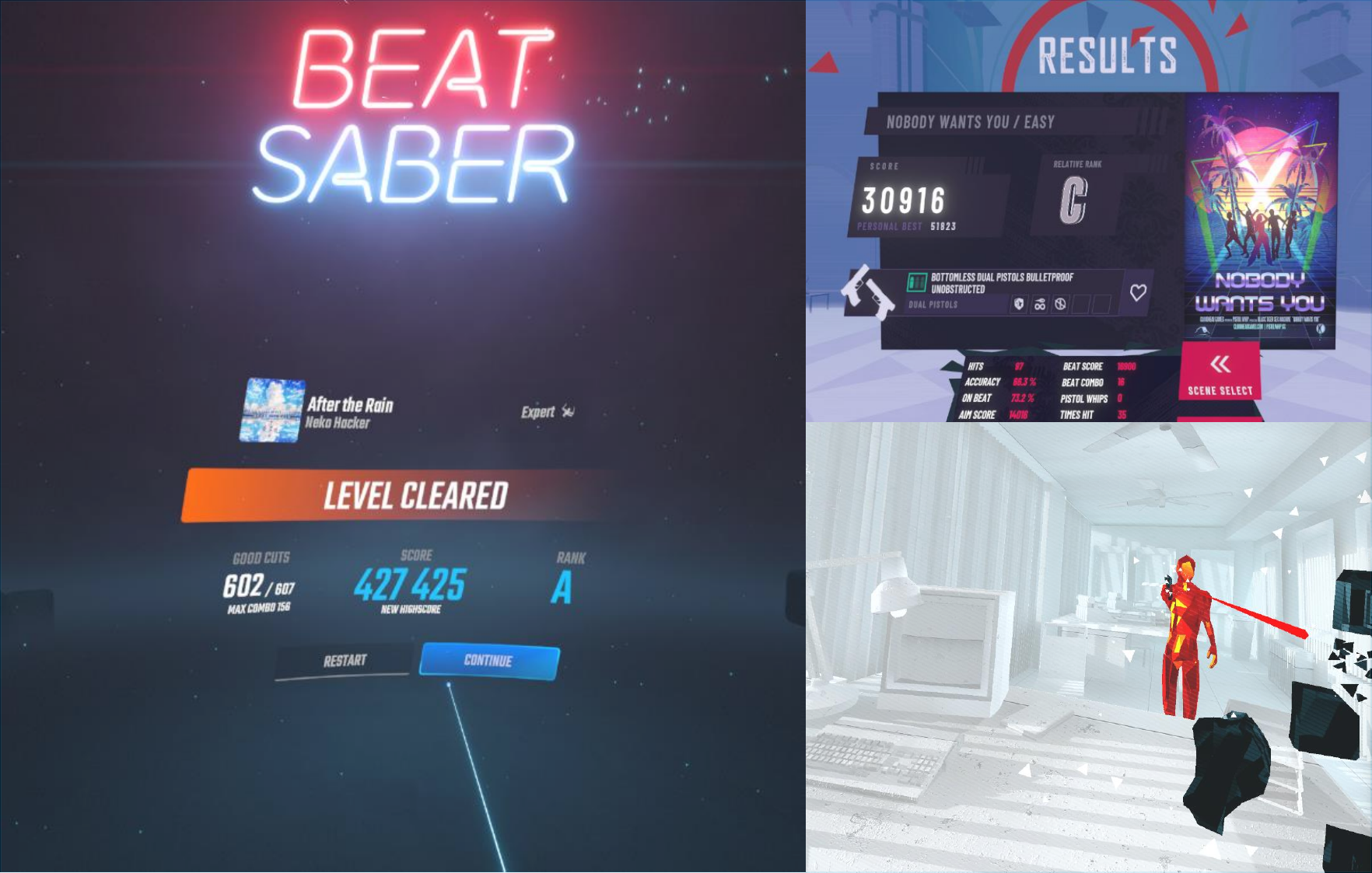}
\caption{Screenshots for each VR title: Beat Saber, Pistol Whip,  SuperHot VR.}
\label{fig:gamescore}
\end{figure}

\noindent\textbf{Performance in Games:}
To assess performance under different data volumes, we began with 1 hour of gameplay demonstration, then 2 hours, and finally 3.5 hours. With only 1 hour of data, the agent could not perform correct actions in game. With 2 hours, it performed reasonable actions such as hitting the correct cubes in Beat Saber and shooting in Pistol Whip. Once trained with 3.5 hours of data, the agent played more like a human player, clearing maps in Beat Saber and combining shooting with head movement in Pistol Whip. At this stage, the evaluation across games shows that in Beat Saber the agent successfully cleared an Expert-level map with Rank A, demonstrating its ability to handle fast-paced rhythm-based interactions. In Pistol Whip, it exhibited human-demonstration-like behaviour with Rank C, including accurate shooting, avoiding threats, and executing local movements consistent with expert play. Screenshots of in-game scores are provided in Figure~\ref{fig:gamescore} for reference. In SuperHot, despite the higher complexity of gameplay mechanics, the agent was able to perform actions such as grabbing items and avoiding incoming threats, and we illustrate the outcome with a representative scene image instead of score screenshots. This lower performance could be partly attributed to the player's lower familiarity with these two games, leading to poorer quality and coverage of demonstration data~\cite{belkhale2023data,simchowitz2025pitfalls}.

\begin{table*}[h]
\centering
\begin{tabular}{|c|c|c|c|c|c|}
\hline
Map Name & Notes/sec & Max Combo (SW) & Max Combo (No SW) & Accuracy (SW, \% Good) & Accuracy (No SW, \% Good) \\
\hline
Gurenge - Joetastic & 3.27 & 72 & 4 & 96 & 32 \\
ALIVE - tukiomoi & 3.74 & 42 & 15 & 85 & 73 \\
Ao no Sumika - Joetastic & 5.35 & 73 & 11 & 87 & 65 \\
Beat Saber - Jaroslav Beck & 5.72 & 53 & 19 & 81 & 56 \\
Sparkling Daydream - Joetastic & 6.90 & 39 & 35 & 84 & 84 \\
\hline
\end{tabular}
\vspace{2pt}
\caption{Comparison of sliding-window vs. no sliding-window performance across different Beat Saber maps. Each entry represents the average across three runs.}
\label{tab:sliding_window}
\end{table*}

\begin{table*}[h]
\centering
\begin{tabular}{|c|c|c|c|c|c|}
\hline
\textbf{Map Name} & \textbf{Notes/sec} & \textbf{Motion Speed} & \textbf{Prediction Entropy} & \textbf{Action Variance} & \textbf{Historical Consistency} \\
\hline
Gurenge -- Joetastic & 3.27 & 96\% / 72 combo & 95\% / 65 combo & 76\% / 17 combo & 96\% / 110 combo \\
ALIVE -- tukiomoi & 3.74 & 85\% / 42 combo & 71\% / 22 combo & 85\% / 31 combo & 40\% / 9 combo \\
Ao no Sumika -- Joetastic & 5.35 & 87\% / 73 combo & 67\% / 25 combo & 84\% / 33 combo & 38\% / 8 combo \\
Beat Saber -- Jaroslav Beck & 5.72 & 81\% / 53 combo & 49\% / 21 combo & 57\% / 13 combo & 53\% / 11 combo \\
Sparkling Daydream -- Joetastic & 6.90 & 84\% / 39 combo & 79\% / 45 combo & 35\% /  6 combo & 39\% / 8 combo \\
\hline
\end{tabular}
\vspace{3pt}
\caption{Comparison of impact of each different signal on sliding-window performance. Each entry represents the average across three runs.}
\label{tab:factors_window}
\end{table*}

\noindent\textbf{Effect of Dynamic Sliding-Window:} 
To assess the impact of adaptive action aggregation, we used \emph{Beat Saber} as a benchmark because its maps naturally vary in tempo and density, forcing the agent to trade off between rapid reaction and smooth control. Table~\ref{tab:sliding_window} reports three complementary metrics: (i) \emph{Notes per second}, capturing the required reaction speed; (ii) \emph{Max combo}, the longest streak of successful hits, reflecting stability and adaptability; and (iii) \emph{Accuracy}, the percentage of good hits, reflecting overall effectiveness.

Across maps of increasing difficulty, the dynamic sliding-window agent consistently achieved longer combos and higher accuracy than the non-adaptive baseline. For example, on Map~2 (5.35 notes/sec), it sustained a maximum combo of 73 versus 11, with a 22\% higher accuracy. Even on the fastest map (6.90 notes/sec), where maintaining both stability and responsiveness is challenging, the adaptive agent preserved comparable accuracy while avoiding breakdowns in combo streaks.

Beyond aggregate results, our analysis of adaptation signals showed that average motion speed over short horizons was the most reliable factor for adjusting the prediction window in Table~\ref{tab:factors_window}. As the table illustrates, other signals, such as prediction entropy, action variance, and historical consistency, can perform well on a specific map, but motion speed consistently yielded higher accuracy and longer combos across all maps, outperforming the rest. This aligns with the task structure: higher hand/controller velocities coincide with denser or more complex game map patterns, where shorter horizons improve reactivity, while slower segments benefit from longer horizons for smoother, less jittery actions. Other factors we tested, were less predictive of performance in this setting, though they may play stronger roles in tasks with different dynamics. These findings suggest that adaptive action aggregation, especially when driven by task-aligned signals like motion speed, enables agents to dynamically balance responsiveness and smoothness.

\section{Discussion}

Our experiments demonstrate that an ACT-based agent with adaptive action aggregation can achieve reliable performance across diverse VR titles, suggesting its promise as a foundation for automated VR game testing. In this section we discuss how the approach may be extended to more genres, and reflect on its current limitations and possible improvements.

\noindent\textbf{Baseline Model Comparison:}
In this paper, we adopt the well-established ACT model and introduce our own adaptations for VR games. In future work, we plan to explore more advanced state-of-the-art models and conduct broader comparisons to further validate and strengthen our findings.

As a preliminary step, we compared our approach against a CNN-MLP baseline. This baseline model is trained to predict the next action from the current observation, using settings similar to those of the ACT model. Although its training loss converged quickly, the baseline struggled to produce coherent actions during actual gameplay. We attribute this failure to its single-step prediction design, which causes the model to drift rapidly into previously unseen states~\cite{simchowitz2025pitfalls}, thereby preventing stable and meaningful behavior in VR environments.

\noindent\textbf{Generalisation to more VR games:} While our evaluation focused on rhythm (Beat Saber), shooter (Pistol Whip), and time-manipulation (Superhot) genres, the methodology could be extended to a broader spectrum of VR games. Open-world exploration games challenge the agent with long-term planning, navigation through large spaces, and coping with partial observability. Narrative-driven experiences demand sensitivity to story cues and choices that may influence future outcomes, while VR training simulators require precise, repeatable behaviours under strict task protocols. These settings emphasise high-level decision making and flexible exploration strategies, rather than just short-term reactivity. To handle such demands, imitation learning could be combined with reinforcement learning in a hybrid framework. Reinforcement learning would enable the agent to learn by trial and error, explore action options beyond those shown in demonstrations, and adapt policies to rare or unexpected states. Over time, this process can make the agent’s behaviours less distinguishable from those of human players, increasing robustness and realism. Such an extension would broaden the applicability of VR agents as automated testers, capable of simulating realistic human gameplay across varied environments.

\noindent\textbf{Limitations of the dynamic sliding-window approach:} The adaptive window mechanism proved effective in rhythm-based games, where rapid reaction and smooth control must be balanced. However, this strategy may not generalise as effectively to slower or more strategic VR genres, where responsiveness is less important than high-level decision making. Our analysis suggested that motion speed was the most reliable adaptation signal in the current setting, but richer features may be required in other contexts. One promising direction is to introduce a lightweight multilayer perceptron (MLP) that predicts the optimal horizon length directly from contextual features such as action variance, prediction entropy, or environment difficulty. Such a learned horizon predictor could provide more flexible adaptation without retraining the core policy.

\noindent\textbf{Incorporating More Scenario Data:}
Current experiments rely primarily on RGB visual observations, which limits the agent’s awareness of the environment. This misses out on important multimodal signals that are critical in VR games, such as spatial audio cues for direction and timing, or semantic representations of objects that inform interaction possibilities. If additional scenario data such as sound streams, object categories, or scene graphs were incorporated, the agent could better align its decisions with human-like perception. For example, sound could improve reaction to off-screen threats, while object representations could enable more context-aware planning and manipulation. Multimodal integration would therefore not only improve performance but also make the agent’s behaviour more natural and adaptable across a wider range of VR scenarios. We plan to adopt the approach described in \cite{wen2024vr} to retrieve these data.

\section{Conclusion}
In this work we introduced VRScout, a deep learning–based agent designed to autonomously play and evaluate VR games in real time. By extending the Action Chunking Transformer with adaptive action aggregation, our system balances reactivity and smoothness, achieving expert-level performance across diverse commercial VR titles while running efficiently on consumer-grade hardware. These results demonstrate the feasibility of using learning-based agents as scalable tools for automated VR testing.

\bibliographystyle{IEEEtran}
\bibliography{ref}

\end{document}